\documentclass[11pt,a4paper]{article}
\usepackage[hyperref]{acl2019}
\usepackage{times}
\usepackage{latexsym}

\usepackage{url}
\usepackage{booktabs}

\usepackage{caption} 
\usepackage{float}

\usepackage{graphicx}
\usepackage{subcaption}
\usepackage{multirow}
\graphicspath{{images/}}

\usepackage{tabularx}

\newcommand\clearrow{\global\let\rowmac\relax}

\aclfinalcopy % Uncomment this line for the final submission
\title{Measuring social bias in knowledge graph embeddings}

\author{Joseph Fisher\thanks{\hspace{5pt}Corresponding author} \\ Amazon Alexa  \\  Cambridge  \\ fshjos@amazon.com \And
        Dave Palfrey \\ Amazon Alexa \\ Cambridge \\ dpalfrey@amazon.co.uk \And
        Christos Christodoulopoulos  \\ Amazon Alexa \\ Cambridge \\ chrchrs@amazon.co.uk
        \AND
        Arpit Mittal \\ Amazon Alexa \\ Cambridge \\ mitarpit@amazon.co.uk
        }

\date{}

\begin{document}
\maketitle

\begin{abstract}

It has recently been shown that word embeddings encode social biases, with a harmful impact on downstream tasks. However, to this point there has been no similar work done in the field of knowledge graph embeddings. We present the first study on social bias in knowledge graph embeddings, and propose a new metric suitable for measuring such bias. We conduct experiments on Wikidata and Freebase, and show that, as with word embeddings, harmful social biases related to professions are encoded in knowledge graph embeddings with respect to gender, religion, ethnicity and nationality. For example, knowledge graph embeddings encode the information that men are more likely to be bankers, and women more likely to be homekeepers. As knowledge graph embeddings become increasingly utilized, we suggest that it is important the existence of such biases are understood and steps taken to mitigate their impact. 

\end{abstract}

\section{Introduction}

Recent work in the word embeddings literature has shown that embeddings encode gender and racial biases, \cite{manprogrammer,caliskan,100years}. These biases can have harmful effects in downstream tasks including coreference resolution, \cite{coref} and machine translation, \cite{translation}, leading to the development of a range of methods to try to mitigate such biases, \cite{manprogrammer, remove}. In an adjacent literature, learning embeddings of knowledge graph entities and relations is becoming an increasingly common first step in utilizing knowledge graphs for a range of tasks, from missing link prediction, \cite{Transe, Complex}, to more recent methods integrating learned embeddings into language models, \cite{Ernie, hillary, knowbert}.

A natural question to ask is ``do knowledge graph embeddings encode social biases in similar fashion to word embeddings"? We show that existing methods for identifying bias in word embeddings are not suitable for knowledge graph embeddings, and present an approach to overcome this using embedding finetuning. We demonstrate (perhaps unsurprisingly) that unequal distributions of people of different genders, ethnicities, religions and nationalities in the Freebase and Wikidata knowledge graphs result in biases related to professions being encoded in knowledge graph embeddings, such as that men are more likely to be bankers and women more likely to be homekeepers. 

Such biases are potentially harmful when knowledge graph embeddings are used in downstream applications. For example, if embeddings are used in a fact checking task\footnote{Where we evaluate the likelihood that a new triple is correct before adding it to a knowledge base.}, they would make it less likely that we accept facts that a female entity is a politician as opposed to a male entity. Alternatively, as knowledge graph embeddings get utilized as input to language models \cite{Ernie, hillary, knowbert}, such biases can impact all downstream NLP tasks, as has been the case with bias in word embeddings.

We begin in Section \ref{background} by providing the interpretation of bias in embeddings used in this paper, before introducing knowledge graph embedding methods, and discussing how the commonly used method of measuring bias in word embeddings is not applicable to knowledge graph embeddings. In Section \ref{method} we present our proposed alternative approach for identifying bias in graph embeddings. Section \ref{results} then presents results for Wikidata with TransE embeddings, and for FB3M with ComplEx embeddings, with further results provided in the Appendix. 

\section{Background} \label{background}

\subsection{Defining bias in embeddings}

Bias can be thought of as ``prejudice in favor or against a person, group, or thing that is considered to be unfair" \cite{IBMBias}. Because definitions of fairness have changed over time, algorithms which are trained on ``real-world" data\footnote{Such as news articles or a knowledge graph} may pick up associations which existed historically (or still exist), but which are considered undesirable. In the word embedding literature, one common idea is to analyse relationships which embeddings encode between professions and gender, race, ethnicity or nationality. We follow this approach in this paper, though note that our method is equally applicable to measuring the encoded relationship between any set of entities in a knowledge graph.\footnote{For example, we could consider the encoded relationship between a person's nationality and their chances of being a CEO etc.}. 

\subsection{Knowledge Graph Embeddings} 

Knowledge graph embeddings are a vector representation of dimension $d$ of all entities and relations in a knowledge graph. To learn these representations, we define a score function $g(.)$ which takes as input the embeddings of a fact in triple form and outputs a score, denoting how likely this triple is to be correct. 

\[s = \textbf{g}(e_1, r_1, e_2)\]

where $e_{1/2}$ are the dimension $d$ embeddings of entities 1/2, and $r_1$ is the dimension $d$ embedding of relation 1. The score function is composed of a transformation, which takes as input one entity embedding and the relation embedding and outputs a vector of the same dimension, and a similarity function, which calculates the similarity or distance between the output of the transformation function and the other entity embedding. 

Many transformation functions have been proposed, including TransE \cite{Transe}, ComplEx \cite{Complex} and RotatE \cite{rotate}. In this paper we use the TransE and ComplEx functions, and the dot product similarity metric, though emphasize that our approach is applicable to any score function. 

\vspace{10pt}

TransE:
\[s = ((e_1 + r_1), \bar{e_2})\]

ComplEx:

\[s = Re(<e_1, r_1, \bar{e_2}>)\]

We use embeddings of dimension 200, and sample 1000 negative triples per positive, by randomly permuting the lhs or rhs entity. We pass the 1000 negatives and single positive through a softmax function, and train using the cross entropy loss.  All training is implemented using the PyTorch-BigGraph library \cite{ptbg}. 

\subsection{Measuring bias in word embeddings}

The most common technique for exposing bias in word embeddings, the ``Word Embedding Association Test" \cite{caliskan}, measures the cosine distance between entity embeddings and the average entity embeddings of two sets of attribute words (where the sets of attribute words could correspond to e.g. male vs. female). They give a range of examples of biases according to this metric, including that science related words are more associated with ``male", and art related words with ``female". In a similar vein, in \cite{manprogrammer}, the authors use the direction between vectors to expose stereotypical analogies, claiming that the direction between man::doctor is analogous to that of woman::nurse. Despite \cite{fair} exposing some technical shortcomings in this approach, it remains the case that distance metrics appear to be appropriate in at least exposing bias in word embeddings, which has then been shown to clearly propagate to downstream tasks, \cite{coref, translation}. 

\begin{figure}[H]
	\centering
	\caption{Unsuitability of distance based metrics for measuring bias in knowledge graph embeddings}
	\includegraphics[width=0.35\textwidth]{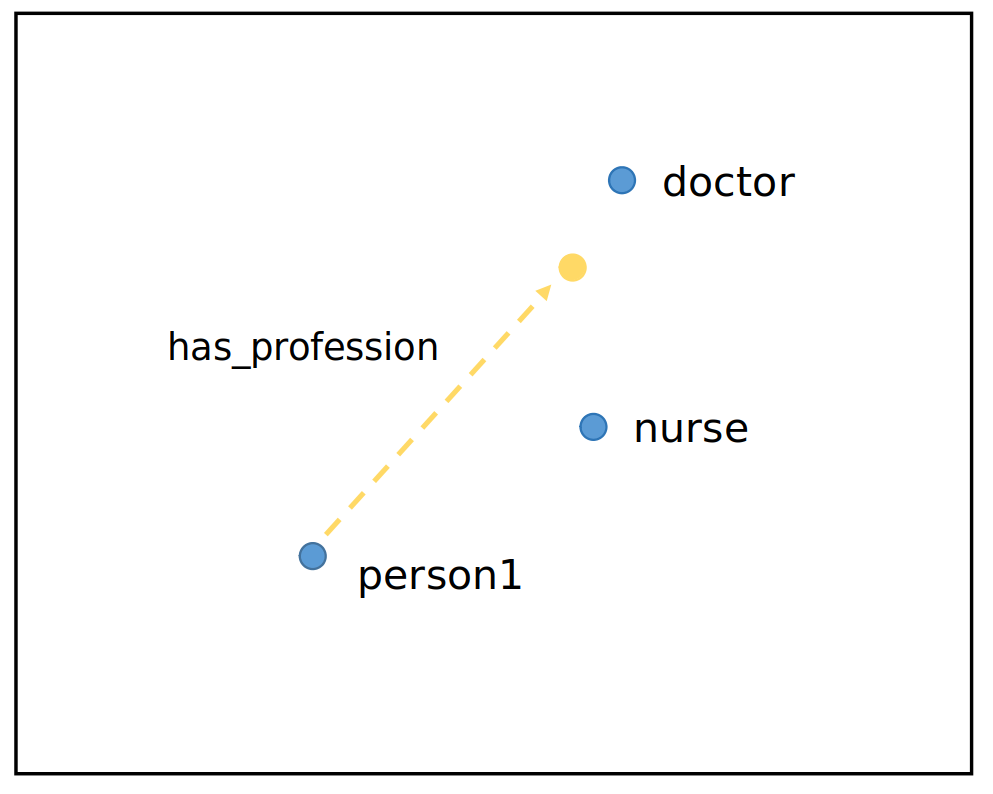}
	\label{fig:distance}
\end{figure}

We suggest that distance-based metrics are not suitable for measuring bias in knowledge graph embeddings. Figure \ref{fig:distance} provides a simple demonstration of the reasoning behind this. Visualizing in a two dimensional space, the embedding of person1 is closer to nurse than to doctor. However, knowledge graph embedding models do not use distance between two entity embeddings when making predictions, but rather the distance between some transformation of one entity embedding with the relation embedding. 

In the simplest case of TransE \cite{Transe} this transformation is a summation, which could result in a vector positioned at the yellow dot in Figure \ref{fig:distance}, when making a prediction of the profession of person1. As the transformation function becomes more complicated, \cite{Complex, rotate} etc., the distance metric becomes increasingly less applicable, as associations in the distance space become less and less correlated with associations in the score function space. 

\section{Method} \label{method}

When presenting measures of bias for knowledge graph embeddings, we define the sensitive attribute we are interested in, denoted $g$, and two alternative values of this attribute, denoted $a$ and $b$. For the purposes of our example we use gender as the sensitive attribute $g$, and male and female as the alternative values $a$ and $b$. We are interested in measuring the influence of the sensitive attribute (gender) on the model's predictions of the likelihood of a person having a profession $p$, defined by the score function:

\[s_{j, p} = \textbf{g}(e_{j}, r_{p}, e_{p})\]

where $e_j$ is the entity embedding of person $j$, $r_p$ is the embedding of the relation corresponding to ``has\_profession", and $e_p$ is the entity embedding of profession $p$. 

\subsection{Comparing pairs of entities}  

Taking into account the discussion of bias in word embeddings, one potential starting point for a bias measure in graph embeddings is to look at the model's scores $s_{j, p}$ for a male entity $j$, and compare them to the scores $s_{i, p}$ for a female entity $i$. For example, we can calculate the scores that the entity ``Barack Obama" has each profession in the knowledge graph, and the scores that the entity ``Michelle Obama" has each profession, and use the difference between the two as a measure of bias, denoted $b_p$:

\[b_p = \textbf{g}(e_{barack}, r_{p}, e_{p}) - \textbf{g}(e_{michelle}, r_{p}, e_{p})\]

\begin{table}[h]
    \centering
    \small
        \caption{Top 20 male professions in Wikidata relative to female using Barack Obama vs. Michelle Obama}
    \label{tab:wikidata_barack_vs_michelle}
    \begin{tabular}{lrrr}
\toprule
                  Profession &  $B_p$ &  $C_{male}$ &  $C_{fem.}$ \\
        \midrule
                zoologist                      &   8.10 &  5355 &   754 \\
                President of the U.S. &  7.71 &     1 &     0 \\
                police officer                 &  6.85 &  2765 &   203 \\
                ornithologist                  &  6.56 &  2283 &   151 \\
                geographer                     &  6.46 &  3922 &   390 \\
                entomologist                   &  5.67 &  4775 &   544 \\
                darts player                   &  5.36 &   681 &    57 \\
                caricaturist                   &  5.11 &  1303 &    63 \\
                child actor                    &  5.03 &   978 &  1074 \\
                theater director               &  5.01 &  6256 &  1563 \\
                biochemist                     &     5.00 &  2136 &   564 \\
                playwright                     &  4.96 &  9293 &  1730 \\
                rikishi                        &  4.88 &   398 &     2 \\
                geologist                      &  4.86 &  4911 &   402 \\
                supervillain                   &  4.85 &    59 &    29 \\
                cartoonist                     &  4.84 &  2380 &   464 \\
                animator                       &  4.78 &  2354 &   410 \\
                rakugoka                       &  4.78 &   181 &    19 \\
                ballet dancer                  &  4.71 &  1126 &  1738 \\
                psychiatrist                   &  4.58 &  3351 &   471 \\
        \bottomrule
\end{tabular}
\end{table}

\begin{table}[h]
    \centering
    \small
        \caption{Top 20 female professions in Wikidata relative to male using Barack Obama vs. Michelle Obama}
    \label{tab:wikidata_michelle_vs_barack}
    \begin{tabular}{lrrr}
     \toprule
       Profession &  $B_p$ &  $C_{fem.}$ &  $C_{male}$ \\
        \midrule
            lawyer                     &  8.66 &   5691 &  55205 \\
            printer                    &  7.91 &     78 &   1785 \\
            typographer                &  7.42 &     19 &    442 \\
            publisher                  &   7.30 &    582 &   6376 \\
            astronomer                 &  6.73 &    565 &   5543 \\
            fashion photogr.       &   6.60 &     49 &    227 \\
            digital artist             &  6.58 &     18 &     57 \\
            visual artist              &  6.51 &   1203 &   2002 \\
            performance artist         &  6.27 &    181 &    203 \\
            architectural photogr. &  6.26 &     16 &     83 \\
            military commander         &  6.02 &      0 &   1077 \\
            photojournalist            &  6.02 &    215 &    918 \\
            Hofmeister                 &   5.90 &     13 &    125 \\
            notary                     &  5.89 &     34 &    874 \\
            attorney at law            &  5.82 &     14 &     89 \\
            editor                     &  5.74 &   1177 &   5488 \\
            count                      &  5.74 &     13 &    102 \\
            advocate                   &   5.70 &    110 &    486 \\
            painter                    &  5.69 &  16514 &  87371 \\
            translator                 &  5.64 &   6298 &  21045 \\
        \bottomrule
\end{tabular}

\end{table}

Table \ref{tab:wikidata_barack_vs_michelle} shows the top 20 ``male" professions in Wikidata using this metric, and Table \ref{tab:wikidata_michelle_vs_barack} the top 20 ``female" professions using the inverse metric. Alongside the score we present the counts of humans in the knowledge graph which have this profession, split by attributes. For example, the top rows of column $C_{male}$ and $C_{fem.}$ in Table \ref{tab:wikidata_barack_vs_michelle} shows that there are 5355 male entities in Wikidata with the profession ``zoologist", and 754 female entities with this profession.\footnote{Whilst for the majority of results tables we limit professions to those with at least 20 occurrences in the knowledge graph to reduce noise, we leave in professions such as ``President of the United States" which have less observations than this but have Barack or Michelle as a left hand entity.}

The top professions, whilst in some cases amusing (``supervillain" and ``ballet dancer" for Obama!) appear uncorrelated with the male/female counts in Wikidata; for example ``military commander" is in the top 20 female professions by this measure whilst having no female observations and 1077 male observations. In general, we find this corresponds to a pattern with knowledge graph embeddings that measures based on a single entities' embeddings are highly noisy, and depending significantly on the particular set of triples which the corresponding entity is included in. To illustrate this, Tables \ref{tab:donald_vs_melania} and \ref{tab:melania_vs_donald} in Appendix \ref{app:trump} show the same measure, but using Donald and Melania Trump. The results have very little overlap with those in Tables \ref{tab:wikidata_barack_vs_michelle} and \ref{tab:wikidata_michelle_vs_barack}, suggesting they depend more on the particular entities chosen than the model's representation of gender relative to professions. 

A potential alternative to this is to average such comparisons across multiple pairs of male/female entities. However, as we know that the distribution of human entities for each profession in real-world knowledge graphs with sensitive attributes is not balanced (there are, as we have seen, many more male ``military commanders" in Wikidata than female), the resulting measure would simply represent whether the model is able to give higher scores to people's correct professions. Instead, we are interested in analysing if the trained embeddings of professions encode the information that they are more male/female, or put another way, if the model is likely to attribute a higher likelihood to an entity having a particular profession purely on the basis of their gender. 

\subsection{Proposed metric}

\begin{figure}[h]
	\centering
	\caption{Finetuning of embedding along gender axis}
	\includegraphics[width=0.37\textwidth]{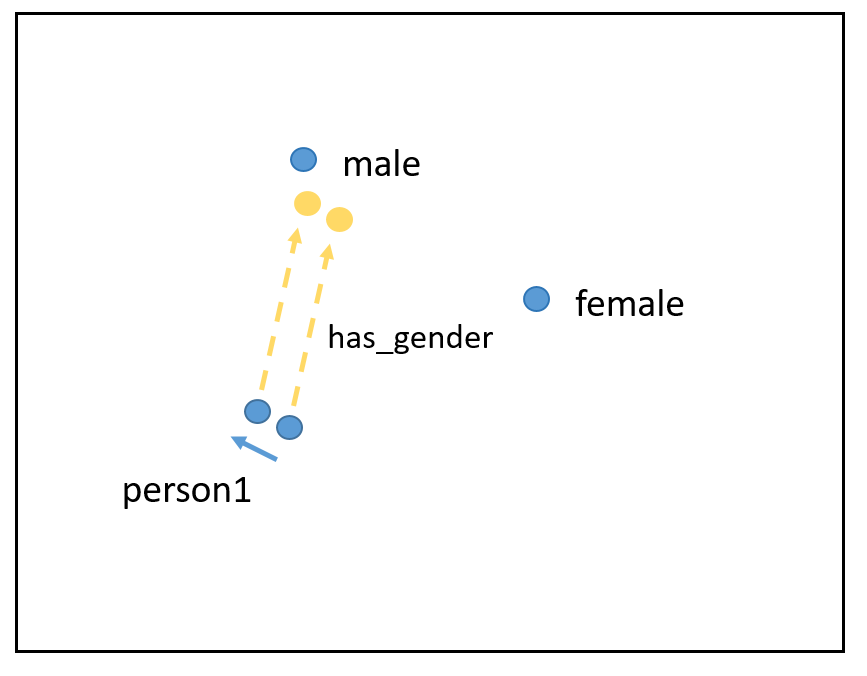}
	\label{fig:pre}
\end{figure}

In light of this discussion, we propose the following metric. We take a trained embedding of a human entity, $j$, denoted $e_j$ and calculate an update to this embedding which increases the score that they have attribute $a$ (male), and decreases the score that they have attribute $b$ (female). In other words, we finetune the embedding to make the person ``more male" according to the model's own encoding of masculinity. This is visualized for TransE in Figure \ref{fig:pre}, where we shift person1's embedding so that the transformation between person1 and the relation has\_gender moves closer to male and away from female. 

Mathematically, we define function $\textbf{m}$ as the difference between the score that person $j$ has sensitive attribute $a$ (male) and that they have sensitive attribute $b$ (female). We then differentiate $\textbf{m}$ wrt the embedding of person $j$, $e_{j}$, and update the embedding to increase this score function. 
\begin{equation}
    \textbf{m}(\theta) = \textbf{g}(e_{j}, r_{s}, e_{a}) - \textbf{g}(e_{j}, r_{s}, e_{b})
\end{equation}

\[e_{j}' = e_{j} + \alpha \frac{\delta \textbf{m}(\theta)}{\delta e_{j}}\]

\vspace{10pt}

where $e_{j}'$ denotes the updated embedding for person $j$, $r_{g}$ the embedding of the sensitive relation $i$ (gender), and $e_{a}$ and $e_b$ the embeddings of attributes $a$ and $b$ (male and female). This is equivalent to providing the model with a batch of two triples, $(e_{j}, r_{g}, e_{a})$ and $(e_{j}, r_{g}, e_{b})$, and taking a step with the basic gradient descent algorithm with learning rate $\alpha$.\footnote{In all experiments in the paper we set $\alpha$ to 0.01. For TransE, the linearity of the score function means the result is independent of $alpha$. For non-linear score functions, the value of $\alpha$ should be kept small to ensure the updated embeddings $e_{j}'$ remain in the proximity of unchanged human embeddings in the knowledge graph.} 

We then analyse the change in the models scores for each profession. That is, we calculate whether, according to the model's score function, making an entity more male increases or decreases the likelihood that they have a particular profession, $p$:

\[\nabla_{p} = \textbf{g}(e_{j}', r_{p}, e_{p}) - \textbf{g}(e_{j}, r_{p}, e_{p})\]

where $e_{p}$ denotes the entity embedding of the profession, $p$. 

\begin{figure}[h]
	\centering
	\caption{Effect of finetuning on scores of professions}
	\includegraphics[width=0.38\textwidth]{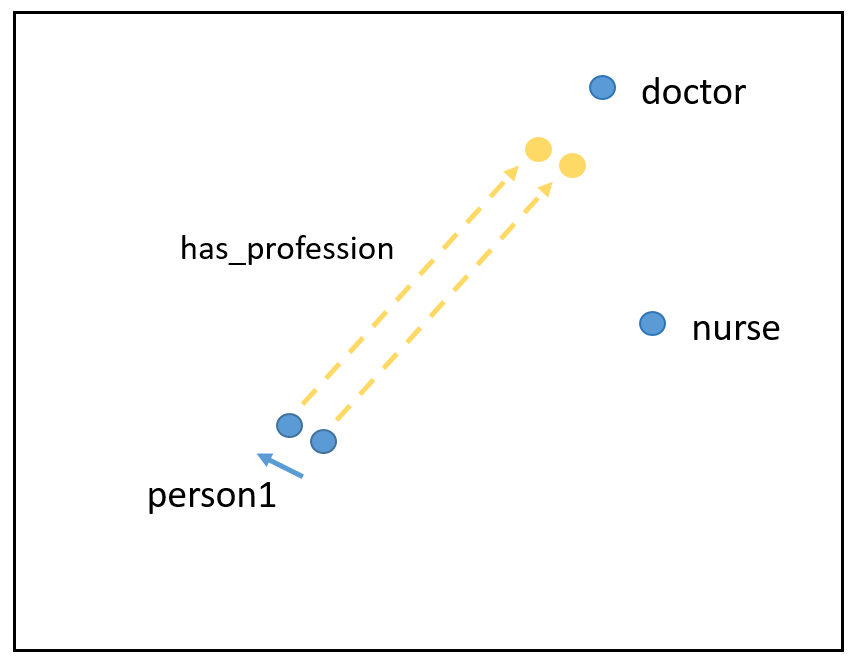}
	\label{fig:post}
\end{figure}

Figure \ref{fig:post} illustrates this. The adjustment to person1's embedding defined in Figure \ref{fig:pre} results in the transformation of person1 and the relation has\_profession moving closer to doctor and further away from nurse. That is, the score g(person1, has\_profession, doctor) has increased, and the score g(person1, has\_profession, nurse) has decreased. In other words, the embeddings in this case encode the bias that doctor is a profession associated with male rather than female entities.

We can then repeat the process for all humans in the knowledge graph and calculate the average changes, giving a bias score $b_p$ for profession $p$:

\[b_p = \frac{1}{J}\sum_{j=1}^{J}\nabla_{p}\]

where J is the number of human entities in the knowledge graph. We calculate this score for each profession, $p = 1,...,P$ and rank the results. 

Importantly, this metric does not involve complementing/altering the training data or training procedure, meaning the results presented below are applicable to knowledge graph embeddings as trained in the standard fashion in the literature. In addition, the measure can be calculated for/averaged across all human entities in the graph, and as such, does not depend on the particular entity/set of entities chosen. 

\section{Results} \label{results}

We provide results in the main paper for Wikidata using TransE \cite{Transe} embeddings, and for FB3M using ComplEx embeddings, providing a demonstration of our method for two transformation functions and two knowledge graphs. Additional results for Wikidata are provided in Appendix \ref{app_wiki} for bias along religious and national lines, and for FB3M in Appendix \ref{app_fb3m} for bias along gender, ethnic, religious and national lines. In all results tables, we limit to professions which have at least 20 observations in the knowledge graph.

\subsection{TransE embeddings (Wikidata)}

\begin{table}[h]
    \centering
    \small
    \caption{Top 20 male professions in Wikidata relative to female using TransE embeddings}
    \label{tab:wikidata_most_male}
    \begin{tabular}{lrrr}
     \toprule
                  Profession &  $B_p$ &  $C_{male}$ &  $C_{fem.}$ \\
        \midrule
                    baritone &  0.132 &           44 &            0 \\
          military commander &  0.128 &         1077 &            0 \\
                      banker &  0.121 &         6664 &          280 \\
               racing driver &  0.106 &         3152 &          139 \\
                    engineer &  0.103 &        27333 &         1124 \\
                    explorer &  0.102 &         5360 &          315 \\
                     luthier &  0.101 &          261 &            0 \\
              chess composer &  0.101 &          614 &            4 \\
          F1 driver &  0.100 &          681 &            3 \\
                     prelate &  0.099 &         1573 &            2 \\
          forestry scientist &  0.097 &          147 &            1 \\
                       count &  0.095 &          102 &           13 \\
             military leader &  0.093 &         5029 &           33 \\
            motorcycle racer &  0.091 &         2855 &           89 \\
                      jockey &  0.091 &         1327 &           89 \\
                      priest &  0.089 &        21781 &          270 \\
                      pastor &  0.088 &         2565 &           85 \\
         structural engineer &  0.088 &          212 &            3 \\
             local historian &  0.088 &          970 &           52 \\
             legal historian &  0.088 &          748 &           49 \\
        \bottomrule
    \end{tabular}
\end{table}

\begin{table}[h]
    \centering
    \small
    \caption{Top 20 female professions in Wikidata relative to male using TransE embeddings}
    \label{tab:wikidata_most_female}
    \begin{tabular}{lrrr}
    \toprule
       Profession &  $B_p$ &  $C_{fem.}$ &  $C_{male}$ \\
        \midrule
                      nun &  0.174 &            1754  & 8\\
                 feminist &  0.145 &           1441  & 26\\
                  soprano &  0.138 &            110  & 2\\
              Suffragette &  0.126 &            1073 & 0\\
            mezzo-soprano &  0.126 &            28 & 0\\
          salonniere &  0.126 &           444 & 16 \\
               homekeeper &  0.120 &        322 & 1 \\
                 princess &  0.118 &        128 & 0 \\
            queen consort &  0.115 &        21 & 0 \\
                 activist &  0.110 &        2102 & 1344 \\
                    nurse &  0.108 &        1896 & 212 \\
         woman of letters &  0.107 &        165 & 10 \\
                   abbess &  0.103 &     98 & 0 \\
               suffragist &  0.101 &        689 & 54 \\
           textile artist &  0.101 &        714 & 195 \\
               prostitute &  0.101 &        195 & 23 \\
                     maid &  0.100 &        51 & 1 \\
         rhythmic gymnast &  0.099 &        915 & 1 \\
                  AV Idol &  0.099 &        2176 & 1 \\
            fashion model &  0.098 &        1670 & 17 \\
        \bottomrule
    \end{tabular}
\end{table}

Tables \ref{tab:wikidata_most_male} and \ref{tab:wikidata_most_female} present the results for gender, with attribute $a$ being male and $b$ female. Whilst the discrepancies in counts are of interest in themselves \cite{manwiki} our main aim in this paper is to show that these differences propagate to the learned embeddings. Table \ref{tab:wikidata_most_male} confirms this; although it includes a number of professions which are potentially male by definition, such as ``baritone",\footnote{The decision over which professions should be allowed to vary with gender/religion etc. is difficult, and dependent on the particular user/application. As such we defer from giving a fixed set of professions for which the likelihoods should be allowed to vary with each sensitive attribute in this paper.} there are also many which we may wish to be neutral, such as ``banker" and ``engineer". Whilst there is a strong correlation between the counts and $B_p$, it is not perfect. For example, there are more male and less female priests than there are bankers, but we get a higher score according to the model for banker than we do priest. The interconnected nature of graphs makes diagnosing the reason for this difficult, but there is clearly a difference in representation of the male entities in the graph who are bankers relatives to priests, which plays out along gender lines. 

Table \ref{tab:wikidata_most_female} presents the most female professions relative to male for Wikidata (i.e. we reverse $a$ and $b$ from Table \ref{tab:wikidata_most_male}). As with the most male case, there are a mixture of professions which are female by definition, such as ``nun", and those which we may wish to be neutral, such as ``nurse" and ``homekeeper". This story is supported by Tables \ref{tab:FB3M_most_male} and \ref{tab:FB3M_most_female} in the Appendix, which give the same results but for the FB3M dataset. 

\begin{table}[h]
    \centering
    \small
    \caption{Top 20 most Jewish professions in Wikidata relative to ethnicity African American with TransE embeddings}
    \label{tab:wikidata_most_jewish}
    \begin{tabular}{lrrr}
    \toprule
          Profession &  Score & $C_{Jew.}$ &  $C_{AfAm.}$ \\
        \midrule
          opinion journalist &  0.217 &            22 &           2 \\
                       rabbi &  0.206 &            71 &           5 \\
            theater director &  0.190 &           9 &            32 \\
                 sociologist &  0.130 &           16 &           40 \\
             literary critic &  0.123 &           34 &           19 \\
                   publisher &  0.122 &           16 &           18 \\
                  translator &  0.112 &            116 &          5 \\
                entrepreneur &  0.108 &           50 &           66 \\
                   economist &  0.104 &           27 &           15 \\
                restaurateur &  0.089 &           1 &            21 \\
         film score composer &  0.088 &           10 &           25 \\
                      editor &  0.087 &           10 &           30 \\
         political scientist &  0.081 &           8 &            13 \\
                    engineer &  0.079 &           27 &           54 \\
                  biographer &  0.078 &           12 &           25 \\
                 stage actor &  0.074 &          50 &           406 \\
                    linguist &  0.073 &            27 &           4 \\
                   historian &  0.072 &           68 &           82 \\
                    inventor &  0.070 &           19 &           58 \\
          computer scientist &  0.065 &           13 &           15 \\
        \bottomrule
    \end{tabular}
\end{table}

\begin{table}[h]
    \centering
    \small
    \caption{Top 20 most African American professions in Wikidata relative to ethnicity Jewish with TransE embeddings}
    \label{tab:wikidata_most_afr}
    \begin{tabular}{lrrr}
     \toprule
               Profession &  Score &   $C_{AfAm.}$&  $C_{Jew.}$ \\
        \midrule
         Canadian football player &  0.217 &   298 &    0 \\
         American football player &  0.180 &  1661 &    1 \\
                       head coach &  0.175 &    41 &    0 \\
                  baseball player &  0.161 &   979 &    0 \\
             mixed martial artist &  0.137 &    60 &    0 \\
                    visual artist &  0.132 &    57 &    1 \\
                           dancer &  0.122 &   186 &    7 \\
            civil rights advocate &  0.121 &    73 &    0 \\
             motivational speaker &  0.114 &    38 &    1 \\
                 basketball coach &  0.107 &   363 &    1 \\
                singer-songwriter &  0.107 &   559 &   12 \\
               pornographic actor &  0.103 &    61 &    9 \\
                            boxer &  0.101 &   149 &    1 \\
                    jazz musician &  0.101 &   698 &    5 \\
                         sprinter &  0.099 &   112 &    1 \\
                 television actor &  0.098 &  1123 &   50 \\
                         academic &  0.098 &    51 &    6 \\
                         minister &  0.097 &    49 &    1 \\
                        guitarist &  0.094 &   255 &    3 \\
                           rapper &  0.094 &   900 &    1 \\
        \bottomrule
    \end{tabular}
\end{table}

We can also calculate biases for other sensitive relations such as ethnicity, religion and nationality. For each of these relations, we choose two attributes to compare, and finetune the embeddings to increase the score of the primary attribute whilst simultaneously reducing the score of the secondary attribute. In Table \ref{tab:wikidata_most_jewish}, we show the professions most associated with the ethnicity ``Jewish" relative to ``African American". As previously, the results include potentially harmful stereotypes, such as the ``economist" and ``entrepreneur" cases. It is interesting that these stereotypes play out in our measure, despite the more balanced nature of the counts \footnote{the balanced counts are themselves due to there being many more entities with ethnicity ``African American" in Wikidata (16280) than ethnicity ``Jewish" (1588).}. In some extreme cases, such as for the profession ``publisher" in Table \ref{tab:wikidata_most_jewish}, the count of people with ``African American" ethnicity (18) is actually greater than the count for people with ``Jewish" ethnicity (16), but the embeddings still encode this as a ``Jewish" profession. Given the interconnected nature of knowledge graphs, it is difficult to precisely diagnose the reason for this, but it is clear that our finetuning based approach is able to identify some biases which would be missed with a simple count-based measure.

\subsection{ComplEx embeddings (FB3M)} \label{complex}

Our method is equally applicable to any transformation function. To demonstrate this, we trained embeddings of the same dimension on the FB3M dataset using the ComplEx transformation \cite{Complex}, and provide the results for gender in Tables \ref{tab:FB3M_most_male_complex} and \ref{tab:FB3M_most_female_complex} below (computational cost prohibited training of ComplEx embeddings on the full Wikidata knowledge graph). 

\begin{table}[h]
    \centering
    \small
    \caption{Top 20 male professions in FB3M relative to female using ComplEx embeddings}
    \label{tab:FB3M_most_male_complex}
    \begin{tabular}{lrrr}
      \toprule
                   Profession &  Score &    $C_{male}$&  $C_{fem.}$ \\
        \midrule
                           /m/0513qg &  0.186 &    160 &    8 \\
                            detective &  0.163 &     27 &    2 \\
                            trumpeter &  0.161 &    346 &    6 \\
                             gangster &  0.146 &     45 &    0 \\
                 private investigator &  0.142 &     18 &    4 \\
         assn. football manager &  0.132 &    587 &    5 \\
                           Trombonist &  0.131 &    196 &    1 \\
                     session musician &  0.130 &    184 &    7 \\
                               sailor &  0.119 &    429 &   23 \\
                            bodyguard &  0.117 &     33 &    2 \\
                           bandleader &  0.115 &    533 &   32 \\
          assn. football player &  0.115 &  13321 &  227 \\
                              samurai &  0.114 &     26 &    0 \\
                       music director &  0.114 &    643 &   29 \\
                   mastering engineer &  0.111 &     33 &    1 \\
                               clergy &  0.107 &     78 &    4 \\
                      baseball umpire &  0.107 &     88 &    0 \\
                                rabbi &  0.105 &    180 &    5 \\
                              Mafioso &  0.103 &     60 &    0 \\
                         statistician &  0.103 &    205 &    3 \\
        \bottomrule
    \end{tabular}
    
\end{table}

\begin{table}[h]
    \centering
    \small
    \caption{Top 20 female professions in FB3M relative to male using ComplEx embeddings}
    \label{tab:FB3M_most_female_complex}
    \begin{tabular}{lrrr}
     \toprule
           Profession &  Score &  $C_{fem.}$&  $C_{male}$ \\
        \midrule
                 gravure idol &  0.210 &   62 & 0 \\
         fitness professional &  0.184 &   24 & 12 \\
           Nude Glamour Model &  0.177 &   511 & 1 \\
                     showgirl &  0.171 &   41 & 0 \\
                          nun &  0.167 &   41 & 0  \\
                    socialite &  0.164 &   81 & 11 \\
                    art model &  0.157 &   22 & 2 \\
             Key Hair Stylist &  0.157 &   43 & 11 \\
           jewellery designer &  0.154 &   39 & 9 \\
                fashion model &  0.153 &   508 & 32 \\
                        nurse &  0.152 &   185 & 20 \\
                   supermodel &  0.151 &   95 & 9 \\
                    Memoirist &  0.148 &   30 & 35 \\
                  Adult model &  0.147 &   24 & 1 \\
                  pin-up girl &  0.146 &   55 & 0 \\
                dialect coach &  0.143 &   14 & 8 \\
                   Prostitute &  0.140 &   63 & 0 \\
             flight attendant &  0.137 &   34 & 3 \\
                ballet dancer &  0.135 &   237 & 104 \\
                  Cheerleader &  0.133 &   20 & 1 \\
        \bottomrule
    \end{tabular}
\end{table}

FB3M contains a different set of professions to Wikidata, but the conclusion that ComplEx embeddings of professions encode potentially harmful social biases remains consistent with the TransE case. It is also clear that our proposed method of measuring this bias is effective at exposing the most gendered professions with the more sophisticated transformation function, where the WEAT would be even less applicable. 

It would be interesting to carry out a comparison of the differences in how bias is encoded for different transformation functions, which we leave to future work, although a qualitative comparison is possible between the FB3M TransE embedding results for gender in Tables \ref{tab:FB3M_most_male} and \ref{tab:FB3M_most_female} (Appendix \ref{app_fb3m}) with the ComplEx embedding results for gender in Tables \ref{tab:FB3M_most_male_complex} and \ref{tab:FB3M_most_female_complex}.

\subsection{Discussion of the binary nature of comparisons}

The metric and results presented are all based around a comparison between \textbf{two} values of a sensitive attribute (male vs. female, African American vs. Jewish etc.). This has two potential problems. Firstly, assigning people a fixed label for some attributes, such as gender, is potentially problematic. However, this is at present a limitation of the structure of knowledge graphs, and as such one which is inherited by any potential measure of bias. Secondly, for all the ``sensitive attributes" discussed in this paper, there are more than two potential attributes, and for some cases, such as nationality, many hundreds. An individual human may have none, one or many of each of these attributes. Whilst the proposed method does not preclude this possibility, we only present the model's representation of the relationship between \textbf{any two} of these alternative values at a time, to keep the comparisons clear. However, if the researcher is interested in analysing the representation of a single nationality vs. all others, this is possible by updating Equation 1 to include the average score of all other nationalities instead of the score of single alternative attribute $b$. This measure will be highly dependent on the choice of nationalities chosen to compare against, \footnote{As well as on the potential weighting required to account for some nationalities having very low counts and noisy embeddings.} and as such, for clarity, we keep the comparisons in this paper between two attributes only.

\section{Summary}

We have presented the first study on social bias in knowledge graph embeddings, and proposed a new metric for measuring such bias. We demonstrated that differences in the distributions of entities in real-world knowledge graphs (there are many more male bankers in Wikidata than female) translate into harmful biases related to professions being encoded in embeddings. Given that knowledge graphs are formed of real-world entities, we cannot simply equalize the counts; it is not possible to correct history by creating female US Presidents, etc. In light of this, we suggest that care is needed when applying knowledge graph embeddings in NLP pipelines, and work needed to develop robust methods to debias such embeddings.

\clearpage

\bibliographystyle{acl_natbib}
\nocite{*}
\bibliography{references}

\clearpage

\appendix

\section{Appendices}
\label{sec:appendix}

\subsection{Wikidata Donald Trumps vs. Melania Trump gender bias scores} \label{app:trump}

\begin{table}[h]
    \centering
    \small
    \caption{Top 20 male professions in Wikidata relative to female using Donald Trump vs. Melania Trump}
    \label{tab:donald_vs_melania}
    \begin{tabular}{lrrr}
     \toprule
                  Profession &  $B_p$ &  $C_{male}$ &  $C_{fem.}$ \\
        \midrule
            business executive   &  5.85 &   3466 &   539 \\
            zoologist            &  5.16 &   5355 &   754 \\
            autobiographer       &   4.7 &   1769 &   784 \\
            entomologist         &  4.42 &   4775 &   544 \\
            marketing            &  4.37 &     14 &     5 \\
            coach                &  4.29 &   3981 &   379 \\
            epidemiologist       &  4.25 &    409 &   174 \\
            ichthyologist        &  4.23 &    840 &   100 \\
            event rider          &  4.04 &    225 &    69 \\
            statistician         &  3.98 &   1625 &   379 \\
            whistleblower        &  3.88 &     33 &     7 \\
            entrepreneur         &  3.75 &  12958 &  1155 \\
            genealogist          &  3.72 &    648 &    42 \\
            film producer        &  3.72 &  16716 &  3468 \\
            character actor      &  3.67 &    254 &    49 \\
            cardiologist         &  3.57 &    596 &    65 \\
            pathologist          &  3.48 &   1069 &   121 \\
            ornithologist        &  3.41 &   2283 &   151 \\
            carcinologist        &   3.4 &    385 &   105 \\
            figure skating coach &  3.39 &    386 &   298 \\
        \bottomrule
    \end{tabular}
\end{table}

\begin{table}[h]
    \centering
    \small
    \caption{Top 20 female professions in Wikidata relative to male using Donald Trump vs. Melania Trump}
    \label{tab:melania_vs_donald}
    \begin{tabular}{lrrr}
       \toprule
       Profession &  $B_p$ &  $C_{fem.}$ &  $C_{male}$ \\
        \midrule
        partisan                 &  10.63 &   178 &  1316 \\
        editor                   &  10.54 &  1177 &  5488 \\
        cultural worker          &  10.09 &    24 &   119 \\
        opinion journalist       &   9.76 &   368 &  3126 \\
        political commissar      &   9.72 &    35 &   438 \\
        agricultural engineer    &   8.92 &    21 &   260 \\
        carver                   &   7.96 &     6 &   122 \\
        slovenist                &    7.4 &    14 &    23 \\
        peasant                  &   7.13 &    63 &   125 \\
        newspaper editor         &   7.08 &   122 &   942 \\
        interpreter              &   7.06 &    69 &   142 \\
        auxiliary bishop         &      7 &     0 &    57 \\
        polonist                 &   6.82 &    31 &    16 \\
        taekwondo athlete        &   6.82 &   910 &  1302 \\
        oenologist               &   6.81 &     9 &    68 \\
        illuminator              &   6.78 &    21 &   387 \\
        Q23957323                &   6.74 &     7 &    22 \\
        disc jockey              &   6.73 &   418 &  3522 \\
        forestry engineer        &    6.7 &     1 &    92 \\
        rural municipality mayor &   6.69 &     4 &    26 \\   
        \bottomrule

    \end{tabular}
\end{table}

\subsection{Wikidata additional results} \label{app_wiki}

We provide a sample of additional results for Wikidata, across ethnicity, religion and nationality. For each case we choose a pair of values (e.g. Catholic and Islam for religion) to compare. 

The picture presented is similar to that in the main paper; the bias measure is highly correlated with the raw counts, with some associations being non-controversial, and others demonstrating potentially harmful stereotypes. Table \ref{tab:wiki_most_uk} is interesting, as the larger number of US entities in Wikidata (390k) relative to UK entities (131k) means the counts are more balanced, and the correlation between counts and bias measure less strong.

\begin{table}[h]
    \centering
    \small
        \caption{Top 20 Catholic professions in Wikidata relative to Islam with TransE embeddings}
    \label{tab:wiki_most_cath}
    \begin{tabular}{lrrr}
\toprule
            Profession &  Score &    $C_{Cat.}$&  $C_{Isl.}$ \\
\midrule
       Catholic priest &  0.361 &  26860 &    0 \\
       Catholic bishop &  0.323 &    189 &    0 \\
                editor &  0.261 &    117 &   18 \\
    literary historian &  0.240 &     25 &    7 \\
      church historian &  0.233 &    198 &    0 \\
            archbishop &  0.226 &    544 &    1 \\
                 canon &  0.223 &    264 &    0 \\
             presbyter &  0.220 &   1099 &    0 \\
    Catholic religious &  0.219 &    310 &    0 \\
         vicar general &  0.217 &    106 &    0 \\
           medievalist &  0.213 &     26 &    0 \\
                bishop &  0.209 &     65 &    0 \\
                 canon &  0.205 &    133 &    0 \\
      auxiliary bishop &  0.204 &     51 &    0 \\
       literary critic &  0.191 &    100 &   55 \\
               brother &  0.190 &    122 &    0 \\
         Prince-Bishop &  0.189 &     89 &    0 \\
        titular bishop &  0.186 &     63 &    0 \\
 classical philologist &  0.185 &     28 &    0 \\
                father &  0.181 &     53 &    0 \\
\bottomrule
\end{tabular}
\end{table}

\begin{table}[h]
    \centering
    \small
        \caption{Top 20 Islamic professions in Wikidata relative to Catholic with TransE embeddings}
    \label{tab:wiki_most_islamic}
\begin{tabular}{lrrr}
\toprule
            Profession &  Score &  $C_{Isl.}$&  $C_{Cat.}$ \\
\midrule
             muhaddith &  0.240 &    284 &    0 \\
                  imam &  0.207 &    173 &    0 \\
            Islamicist &  0.204 &    57 & 5 \\
                 faqih &  0.194 &    317 &    0\\
             Alim &  0.181 &    94 &    0\\
                 mufti &  0.148 &    48 &    0\\
       Qari' &  0.146 &    28 &    0\\
              mufassir &  0.127 &    114 &    0\\
                  qadi &  0.127 &    80 &    0\\
 human rights activist &  0.125 &   59 & 42\\
       record producer &  0.122 &    47 & 8 \\
      religious leader &  0.100 &    19 & 8 \\
             presenter &  0.093 &    30 & 5 \\
               Akhoond &  0.090 &    36 &    0\\
                 model &  0.088 &   240 & 37 \\
            songwriter &  0.081 &   112 & 24 \\
                  Sufi &  0.073 &    23 &    0\\
                mystic &  0.066 &   77 & 21 \\
             Terrorist &  0.066 &    37 & 1 \\
               blogger &  0.065 &   17 & 16 \\
\bottomrule
\end{tabular}

\end{table}

\begin{table}[H]
    \centering
    \small
        \caption{Top 20 nationality ``United Kingdom" professions in Wikidata relative to nationality ``United States" using TransE embeddings}
    \label{tab:wiki_most_uk}
    \begin{tabular}{lrrr}
\toprule
                   Profession &  Score &   $C_{UK}$&   $C_{US}$ \\
\midrule
                civil servant &  0.100 &   150 &   226 \\
            stand-up comedian &  0.095 &   107 &   189 \\
                     comedian &  0.084 &   939 &   829 \\
                    life peer &  0.081 &     1 &    37 \\
                    barrister &  0.080 &     5 &   260 \\
                 bowls player &  0.077 &     2 &   163 \\
       colonial administrator &  0.066 &     5 &    31 \\
           rugby union player &  0.063 &   195 &  2554 \\
                     diplomat &  0.063 &  2254 &  1093 \\
         television presenter &  0.063 &   786 &  1848 \\
                    guitarist &  0.061 &  4049 &  1646 \\
                   agronomist &  0.061 &    26 &     7 \\
                    solicitor &  0.057 &    10 &   106 \\
             fashion designer &  0.055 &   437 &   185 \\
 association football referee &  0.055 &    45 &   159 \\
                 college head &  0.054 &     2 &    24 \\
                    scientist &  0.054 &   881 &   169 \\
                       docent &  0.053 &    21 &    13 \\
                  mountaineer &  0.053 &   211 &   129 \\
                  medievalist &  0.052 &    57 &    61 \\
\bottomrule
\end{tabular}
\end{table}

\begin{table}[H]
    \centering
    \small
        \caption{Top 20 nationality ``United States" professions in Wikidata relative to nationality ``United Kingdom" using TransE embeddings}
    \label{tab:wiki_most_us}
\begin{tabular}{lrrr}
\toprule
               Profession &  Score &   $C_{US}$&  $C_{UK}$ \\
\midrule
    professional wrestler &  0.132 &  1790 &  150 \\
         amateur wrestler &  0.122 &   844 &  162 \\
 Canadian football player &  0.106 &  2163 &    1 \\
             sportswriter &  0.105 &   199 &    0 \\
       pornographic actor &  0.103 &  1800 &   99 \\
                   dancer &  0.102 &  1283 &  163 \\
         baseball manager &  0.097 &   146 &    0 \\
                  manager &  0.097 &   129 &    6 \\
    real estate developer &  0.097 &    28 &    0 \\
                 aikidoka &  0.095 &    29 &    0 \\
    civil rights advocate &  0.095 &    85 &    0 \\
             tribal chief &  0.094 &    42 &    1 \\
                   jockey &  0.092 &   309 &   46 \\
                   pastor &  0.090 &   239 &   22 \\
      landscape architect &  0.089 &   251 &   30 \\
         Playboy Playmate &  0.087 &   317 &    6 \\
             abolitionist &  0.087 &    81 &    2 \\
            urban planner &  0.085 &    74 &   31 \\
     video game developer &  0.084 &    75 &   11 \\
                  gymnast &  0.083 &   122 &   17 \\
\bottomrule
\end{tabular}
\end{table}

\subsection{FB3M results} \label{app_fb3m}

For comparison, we train TransE embeddings on FB3M of the same dimension, and present the corresponding results tables for gender, religion, ethnicity and nationality. The distribution of entities in FB3M is significantly different to that in Wikidata, resulting in a variety of different professions entering the top twenty counts. However, the broad conclusion is similar; the embeddings encode common and potentially harmful stereotypes related to professions. 

\begin{table}[h]
    \centering
    \small
        \caption{Top 20 male professions in FB3M relative to female using TransE embeddings}
    \label{tab:FB3M_most_male}

   \begin{tabular}{lrrr}
    \toprule
              Profession &  Score &  $C_{male}$ &  $C_{fem.}$ \\
\midrule
         baseball umpire &  0.120 &           88 &            0 \\
      Holy Roman Emperor &  0.119 &           23 &            0 \\
          Opera composer &  0.115 &           77 &            0 \\
       Lighting Director &  0.109 &           31 &            2 \\
               surveying &  0.108 &           59 &            0 \\
                arranger &  0.108 &           21 &            0 \\
                  jockey &  0.103 &          124 &            2 \\
              impresario &  0.103 &           79 &            1 \\
             electrician &  0.102 &           43 &            0 \\
   Nordic combined skier &  0.102 &           65 &            0 \\
 Visual Effects Animator &  0.098 &           27 &            2 \\
               Keytarist &  0.097 &           35 &            3 \\
              Trombonist &  0.097 &          196 &            1 \\
                 Mafioso &  0.097 &           60 &            0 \\
                  Pirate &  0.097 &           34 &            1 \\
     electronic musician &  0.097 &           79 &            2 \\
            statistician &  0.096 &          205 &            3 \\
    military engineering &  0.096 &           21 &            0 \\
                chaplain &  0.096 &           71 &            0 \\
        SEO Professional &  0.095 &           99 &            5 \\
\bottomrule
\end{tabular}
\end{table}

\begin{table}[h]
    \centering
    \small
        \caption{Top 20 female professions in FB3M relative to male using TransE embeddings}
    \label{tab:FB3M_most_female}
    \begin{tabular}{lrrr}
\toprule
             Profession &  Score &  $C_{fem.}$&  $C_{male}$ \\
\midrule
           gravure idol &  0.091 &            0 &           62 \\
     Nude Glamour Model &  0.081 &            1 &          511 \\
                  nurse &  0.075 &           20 &          185 \\
          fashion model &  0.067 &           32 &          508 \\
            pin-up girl &  0.060 &            0 &           55 \\
              socialite &  0.058 &           11 &           81 \\
                  model &  0.057 &         1354 &         4680 \\
              housework &  0.056 &            0 &           38 \\
           Hair Stylist &  0.054 &          109 &          307 \\
               stripper &  0.052 &            6 &           52 \\
          ballet dancer &  0.050 &          104 &          237 \\
             Prostitute &  0.047 &            0 &           63 \\
       Key Hair Stylist &  0.047 &           11 &           43 \\
             supermodel &  0.046 &            9 &           95 \\
               showgirl &  0.046 &            0 &           41 \\
      Key Makeup Artist &  0.044 &            9 &           29 \\
 Hair and Makeup Artist &  0.042 &            4 &           24 \\
              secretary &  0.041 &           11 &           43 \\
       registered nurse &  0.041 &            6 &           22 \\
            Adult model &  0.040 &            1 &           24 \\
\bottomrule
\end{tabular}
\end{table}

\begin{table}[h]
    \centering
    \small
        \caption{Top 20 ethnicity Jewish professions in FB3M relative to ethnicity African American with TransE embeddings}
    \label{tab:fb_most_jewish}
    \begin{tabular}{lrrr}
\toprule
           Profession &  Score &  $C_{Jew.}$&  $C_{AfAm.}$ \\
\midrule
                rabbi &  0.098 &    0 &   32 \\
               banker &  0.081 &    2 &   27 \\
            economist &  0.066 &    9 &   42 \\
       Talk show host &  0.052 &   18 &   20 \\
            scientist &  0.051 &    8 &  171 \\
          philosopher &  0.050 &   10 &   92 \\
           playwright &  0.050 &   72 &   80 \\
            physicist &  0.049 &    5 &   84 \\
  film score composer &  0.048 &  106 &  100 \\
        mathematician &  0.048 &    5 &   86 \\
  political scientist &  0.046 &    4 &   17 \\
  television director &  0.046 &   85 &  108 \\
  theatrical producer &  0.044 &    7 &   15 \\
       businessperson &  0.043 &  133 &  253 \\
      patent inventor &  0.042 &   12 &   19 \\
            historian &  0.040 &   35 &   72 \\
   political activist &  0.040 &   11 &    9 \\
 music video director &  0.039 &   14 &    7 \\
           journalist &  0.039 &  161 &  228 \\
             lyricist &  0.036 &   29 &   34 \\
\bottomrule
\end{tabular}
\end{table}

\begin{table}[h]
    \centering
    \small
        \caption{Top 20 ethnicity African American professions in FB3M relative to ethnicity Jewish with TransE embeddings}
    \label{tab:fb_most_afa}
    \begin{tabular}{lrrr}
\toprule
               Profession &  Score &   $C_{AfAm.}$&  $C_{Jew.}$ \\
\midrule
        basketball player &  0.096 &  1489 &    6 \\
                 minister &  0.073 &    23 &    0 \\
                   pastor &  0.055 &    26 &    1 \\
 American football player &  0.054 &   525 &    8 \\
                   rapper &  0.050 &   337 &   11 \\
    professional wrestler &  0.048 &    63 &    9 \\
                    coach &  0.047 &   225 &    7 \\
         basketball coach &  0.042 &    55 &    2 \\
       sports commentator &  0.037 &    23 &    4 \\
               /m/02h669\_ &  0.034 &    46 &   11 \\
              keyboardist &  0.033 &    30 &   11 \\
               bandleader &  0.032 &    68 &    5 \\
                trumpeter &  0.029 &    26 &    0 \\
                  drummer &  0.028 &    44 &   10 \\
                 musician &  0.024 &  1171 &  164 \\
        radio personality &  0.023 &    31 &   25 \\
             Jazz Pianist &  0.023 &    51 &    4 \\
                    model &  0.021 &   147 &   56 \\
           police officer &  0.020 &    18 &    2 \\
              film editor &  0.019 &    19 &   19 \\
\bottomrule
\end{tabular}
\end{table}

\begin{table}[h]
    \centering
    \small
        \caption{Top 20 Catholic professions in FB3M relative to Islam with TransE embeddings}
    \label{tab:fb_most_cath}
    \begin{tabular}{lrrr}
\toprule
               Profession &  Score &  $C_{Cat.}$&  $C_{Isl.}$ \\
\midrule
                   priest &  0.064 &  106 &    0 \\
            visual artist &  0.052 &   44 &    4 \\
       Holy Roman Emperor &  0.052 &   20 &    0 \\
              voice actor &  0.045 &  195 &   11 \\
               playwright &  0.044 &   56 &    6 \\
               theologian &  0.043 &   25 &    5 \\
                 essayist &  0.037 &   26 &    2 \\
                   lawyer &  0.036 &  701 &   68 \\
                barrister &  0.035 &   28 &    9 \\
                 cardinal &  0.034 &   20 &    0 \\
      television director &  0.033 &   66 &   11 \\
          attorney at law &  0.031 &   19 &    1 \\
                  painter &  0.029 &   37 &    6 \\
                  teacher &  0.023 &  149 &   39 \\
                 diplomat &  0.021 &   83 &   35 \\
 American football player &  0.021 &   43 &    3 \\
                   critic &  0.019 &   22 &    2 \\
      television producer &  0.018 &  181 &   23 \\
         fashion designer &  0.018 &   27 &    4 \\
          baseball player &  0.016 &   39 &    1 \\
\bottomrule
\end{tabular}

\end{table}

\begin{table}[h]
    \centering
    \small
        \caption{Top 20 Islamic professions in FB3M relative to Catholic with TransE embeddings}
    \label{tab:fb_most_islamic}
    \begin{tabular}{lrrr}
\toprule
                  Profession &  Score &  $C_{Isl.}$&  $C_{Cat.}$ \\
\midrule
                     warlord &  0.097 &   21 & 0 \\
                   scientist &  0.074 &   39 & 32\\
                      rapper &  0.062 &   49 & 12 \\
                    engineer &  0.045 &   24 & 42 \\
           singer-songwriter &  0.041 &   23 & 71 \\
                  astronomer &  0.040 &   16 & 11\\
           basketball player &  0.038 &   13 & 19 \\
                      singer &  0.035 &   114 & 250 \\
             record producer &  0.035 &   40 & 49 \\
                   professor &  0.033 &   43 & 90 \\
                      editor &  0.032 &   11 & 37 \\
                    lyricist &  0.031 &   13 & 10 \\
               film director &  0.030 &   63 & 148 \\
         film score composer &  0.029 &   29 & 36 \\
            military officer &  0.028 &   15 & 45 \\
                      pundit &  0.027 &   3 & 35 \\
                    comedian &  0.026 &   25 & 127 \\
 association football player &  0.024 &   67 & 58 \\
                 philosopher &  0.024 &   57 & 134 \\
             Social activist &  0.022 &   9 & 15 \\
\bottomrule
\end{tabular}
\end{table}

\begin{table}[h]
    \centering
    \small
        \caption{Top 20 nationality ``United Kingdom" professions in FB3M relative to nationality ``United States" using TransE embeddings}
    \label{tab:fb_most_uk}
    \begin{tabular}{lrrr}
\toprule
           Profession &  Score &  $C_{UK}$&   $C_{US}$ \\
\midrule
            barrister &  0.074 &   142 & 3 \\
            solicitor &  0.058 &   33 & 5 \\
               curler &  0.044 &   11 & 14 \\
  field hockey player &  0.042 &   16 & 16 \\
          broadcaster &  0.041 &   66 & 57 \\
       radio producer &  0.040 &   21 & 33 \\
 television presenter &  0.040 &   877 & 949 \\
            Zoologist &  0.038 &   8 & 12 \\
             Explorer &  0.036 &   25 & 22 \\
                Rower &  0.035 &   43 & 47 \\
           Equestrian &  0.035 &   16 & 14 \\
            cricketer &  0.035 &   86 & 7 \\
           geneticist &  0.032 &   6 & 25 \\
    Radio Broadcaster &  0.031 &   9 & 14 \\
        Civil servant &  0.031 &   11 & 18 \\
              soldier &  0.030 &   1170 & 637 \\
        art historian &  0.030 &   17 & 41 \\
             botanist &  0.030 &   52 & 87 \\
     Business magnate &  0.029 &   13 & 36 \\
  Cross-country skier &  0.028 &   3 & 21\\
\bottomrule
\end{tabular}

\end{table}

\begin{table}[h]
    \centering
    \small
        \caption{Top 20 nationality ``United States" professions in FB3M relative to nationality ``United Kingdom" using TransE embeddings}
    \label{tab:fb_most_us}
    \begin{tabular}{lrrr}
\toprule
                  Profession &  Score &   $C_{US}$&  $C_{UK}$ \\
\midrule
            basketball coach &  0.050 &   415 &    0 \\
              Talk show host &  0.045 &   161 &    6 \\
               Televangelist &  0.043 &    70 &    0 \\
     law enforcement officer &  0.042 &    20 &    0 \\
                  test pilot &  0.042 &    20 &    0 \\
                ADR Director &  0.042 &    41 &    1 \\
        Vaudeville Performer &  0.039 &    83 &    4 \\
                     veteran &  0.039 &    20 &    1 \\
    American football player &  0.038 &  7405 &    3 \\
                     sheriff &  0.037 &    30 &    0 \\
                     Mafioso &  0.036 &    57 &    0 \\
                      cowboy &  0.035 &    26 &    3 \\
              Football Coach &  0.035 &   394 &    1 \\
              news presenter &  0.034 &   241 &    4 \\
 Certified Public Accountant &  0.033 &    33 &    0 \\
            TV Meteorologist &  0.033 &    45 &    0 \\
       motivational speaking &  0.032 &   102 &    6 \\
              police officer &  0.031 &   151 &   10 \\
              Game Show Host &  0.031 &    60 &    2 \\
             attorney at law &  0.030 &    83 &    0 \\
\bottomrule
\end{tabular}

\end{table}

% \clearpage
% \section{Supplemental Material}
% \label{sec:supplemental}

\end{document}